\documentclass{article}

\PassOptionsToPackage{numbers,compress}{natbib}
\usepackage[final,main]{neurips_2026}

\usepackage[utf8]{inputenc}
\usepackage[T1]{fontenc}
\usepackage{hyperref}
\usepackage{url}
\usepackage{booktabs}
\usepackage{amsfonts}
\usepackage{nicefrac}
\usepackage{microtype}
\usepackage{xcolor}
\usepackage{amsmath}
\usepackage{graphicx}
\usepackage{amssymb}
\usepackage{enumitem}

\title{Post-Training is About States, Not Tokens:\\
A State Distribution View of SFT, RL, and On-Policy Distillation}

\author{%
  Dong Nie\\
  Independent Researcher\\
  \texttt{dongnie@cs.unc.edu} \\
}

\begin{document}

\maketitle

\begin{abstract}
Large language model post-training methods such as supervised fine-tuning (SFT),
reinforcement learning (RL), and distillation are often analyzed through their
loss functions: maximum likelihood, policy gradients, forward KL, reverse KL, or
related objective-level variants. We study a complementary factor: the
\emph{state distribution} on which supervision is applied. For an autoregressive
policy, a state is a prompt plus generated prefix. SFT trains on fixed dataset
states, while RL and on-policy distillation (OPD) train on states induced by the
current learner.

We formalize post-training as state-distribution shaping and run a controlled
small-scale study using Qwen3-0.6B-Base on GSM8K, with TruthfulQA and MMLU as
retention evaluations. Our results show three phenomena. First, a mild SFT run
improves GSM8K with little forgetting, while a stress SFT run causes substantial
retention loss. Second, OPD from a degraded SFT teacher surpasses that teacher
on GSM8K, TruthfulQA, and MMLU, despite using the teacher as its only
supervision source. Third, a lightweight on-policy RL run improves GSM8K while
preserving retention. These results support a state-centric view of
post-training: the source and locality of training states can be as important as
the form of the supervision signal. Code can be available at https://github.com/ginobilinie/unifyPostTraining.
\end{abstract}

\section{Introduction}

Post-training is the stage at which a pretrained language model becomes useful
for a particular set of human-facing behaviors. Supervised fine-tuning (SFT)
teaches the model to imitate demonstrations, reinforcement learning (RL)
optimizes sampled model outputs against rewards, and distillation transfers
behavior from a teacher model to a student. These methods are not minor
implementation details: they determine whether a model follows instructions,
solves reasoning problems, refuses unsafe requests, preserves factual knowledge,
and remains robust outside the narrow distribution of its training data.

Despite this practical importance, several basic post-training phenomena remain
awkward to explain with the usual vocabulary. SFT is simple and data-efficient,
yet it can cause catastrophic forgetting or brittle behavior under aggressive
specialization. RL is often optimized with sparse or noisy rewards, yet it can
produce surprisingly stable improvements. Distillation is usually understood as
copying a teacher, yet students sometimes match or outperform teachers. These
observations raise a shared question: what property of the training process
determines whether a post-trained model improves locally or drifts destructively?

The standard answer focuses on objectives. SFT is maximum likelihood on
demonstrations; RL is reward maximization with policy-gradient-style updates;
distillation is a divergence between teacher and student token distributions.
Objective-level analysis is indispensable, but it hides another axis of
variation: \emph{where} the objective is applied. In an autoregressive model, a
token prediction is always conditioned on a state, namely the prompt together
with the generated prefix. Thus two methods with similar token-level signals can
produce different outcomes if they train on different state distributions.

This distinction is especially sharp for SFT, RL, and on-policy distillation
(OPD). SFT applies dense supervision on fixed dataset prefixes. Those prefixes
may not be states that the current model would visit, especially after the model
starts making its own errors. RL, by contrast, samples trajectories from the
current policy and applies reward-derived updates on states the model actually
visits. OPD separates the state source from the signal source: the student
samples the states, while the teacher supplies local guidance. From this
perspective, OPD is closer to RL than to offline distillation, even when its loss
is written as a supervised teacher-student objective.

We argue that this state-source distinction helps explain both stability and
teacher-student reversal. If a teacher has poor global behavior because it
visits undesirable trajectories, a student need not inherit all of those
failures when supervision is queried on the student's own states. Similarly, RL
can be stable not only because of KL penalties or conservative objectives, but
because its updates are naturally localized to the learner's current
trajectories. The core issue is not merely whether the signal is a label, a
reward, or a logit distribution; it is the interaction between that signal and
the state distribution on which it is applied.

We investigate the following thesis:
\begin{quote}
Post-training behavior is governed by the interaction between supervision signal
and training state distribution; on-policy state supervision can preserve
locality and allow students to improve beyond degraded teachers.
\end{quote}

We test this thesis in a controlled single-GPU setting using Qwen3-0.6B-Base.
GSM8K is used as the target task, while TruthfulQA and MMLU measure retention.
The experiments are intentionally small, but they expose the relevant contrasts.
A mild SFT run improves GSM8K with almost no forgetting, showing that SFT is not
inherently destructive. A stress SFT run, however, produces a degraded teacher:
it lowers both target accuracy and retention. OPD from this degraded teacher
then surpasses the teacher on GSM8K, TruthfulQA, and MMLU. Finally, a lightweight
on-policy RL run improves GSM8K while preserving retention. These results do not
support a simplistic claim that one scalar drift metric fully explains
forgetting. Instead, they support a more precise state-centric claim: the source,
locality, and learner-dependence of training states are central to post-training
behavior.

We make three contributions:
\begin{enumerate}[leftmargin=*]
    \item We formulate SFT, RL, and OPD under a common state-distribution view of
    autoregressive post-training.
    \item We implement a controlled single-GPU experimental pipeline measuring
    target accuracy, retention, forgetting, and rollout-state drift.
    \item We provide evidence that OPD can outperform a degraded SFT teacher and
    that on-policy RL improves target performance with little retention loss.
\end{enumerate}

\section{State Distribution View}

\subsection{Autoregressive States}

An autoregressive language model defines a policy
\begin{equation}
    \pi_\theta(y_t \mid x, y_{<t}).
\end{equation}
We call
\begin{equation}
    s_t = (x, y_{<t})
\end{equation}
the \emph{state}. The next token $y_t$ is the action, and a generated answer is a
trajectory through states. Let $d^\pi(s)$ denote the state visitation
distribution induced by policy $\pi$ on a prompt distribution.

This definition is deliberately simple. A state is not a hidden activation, a
training example, or a single token position in isolation. It is the full
conditioning context on which the next-token policy acts. In an LLM, the same
target token can have very different meaning depending on the prefix state in
which it appears. For example, predicting a number after a correct chain of
reasoning and predicting the same number after an inconsistent chain of
reasoning are different policy updates because they touch different conditional
contexts.

Given a prompt distribution $\rho(x)$, a policy induces a trajectory
\begin{equation}
    \tau = (s_1,y_1,s_2,y_2,\ldots,s_T,y_T),
\end{equation}
where $s_{t+1}=(x,y_{\leq t})$. The induced state distribution can be written
informally as
\begin{equation}
    d^\pi(s) =
    \mathbb{E}_{x\sim \rho,\, y_{<t}\sim \pi}
    \left[
    \frac{1}{T}\sum_{t=1}^{T}\mathbf{1}\{s_t=s\}
    \right].
\end{equation}
In practice this distribution is never observed exactly; we approximate it with
rollouts and sampled prefixes. The important point is that $d^\pi$ changes when
the policy changes. Therefore post-training does not only change token
probabilities at fixed contexts; it changes the future contexts the model will
visit.

\subsection{Two Axes: State Source and Signal Source}

A post-training method can be decomposed into two choices. The first is the
\emph{state source}: where the contexts $s$ come from. The second is the
\emph{signal source}: what target, reward, or distribution is used to update the
policy at those contexts. Many discussions collapse these axes into the name of
the algorithm. For example, SFT means dataset states plus human or synthetic
answers; RL means policy states plus rewards; distillation means teacher
signals, but the state source may be teacher rollouts, dataset prompts, or
student rollouts.

Separating these axes clarifies why methods with similar-looking objectives can
behave differently. A teacher-student KL on teacher-generated prefixes is an
offline imitation problem. The same teacher queried on student-generated
prefixes is an on-policy correction problem. Likewise, token cross entropy on
gold prefixes is not equivalent to token cross entropy on learner-induced
prefixes, even if both losses are supervised. The state source determines the
region of policy space in which the signal is applied.

\subsection{SFT: Off-Policy State Fitting}

SFT minimizes token loss on dataset states:
\begin{equation}
    \mathcal{L}_{\mathrm{SFT}}(\theta)
    =
    -\mathbb{E}_{s \sim d_{\mathrm{data}}}
    \log \pi_\theta(y^\star \mid s).
\end{equation}
The supervision is dense, but the states are off-policy. If dataset trajectories
are far from the model's own rollouts, updates may affect behavior in regions
that the model cannot reliably reach or recover from. This is the familiar
exposure-bias problem recast as a state-distribution mismatch.

The off-policy nature of SFT has two consequences. First, it can be very
efficient when the dataset states are close to useful model states: every token
provides a dense learning signal, and the model can acquire a capability without
expensive sampling. This is what we observe in our mild SFT run. Second, SFT can
be brittle when the dataset trajectory distribution is narrow or when training
pressure is too high. The model is repeatedly pushed toward behavior that is
valid on demonstration prefixes, but it is not explicitly trained to recover
from its own prefixes. Under aggressive specialization, this can modify the
policy in ways that harm unrelated capabilities and even harm the target task.

In this view, catastrophic forgetting is not caused by maximum likelihood alone.
It arises when dense off-policy updates move the policy in regions that interact
poorly with the model's own future state distribution. Thus the relevant
question is not simply whether SFT uses forward KL or token cross entropy, but
whether the dataset states are compatible with the learner's induced states.

\subsection{RL: On-Policy Local Improvement}

RL samples trajectories from the current policy and updates the model using
rewards:
\begin{equation}
    \max_\theta \mathbb{E}_{s \sim d^{\pi_\theta}, y \sim \pi_\theta(\cdot|s)}
    [r(s,y)].
\end{equation}
The reward may be sparse, but the states are on-policy. This makes RL a local
improvement procedure: it modifies behavior where the current model actually
visits.

The on-policy property gives RL a different failure mode from SFT. RL can be
sample-inefficient because rewards may be sparse and high-variance, but its
updates are grounded in the learner's own rollouts. If a model frequently enters
a certain reasoning pattern, reward feedback is applied there. If it never
visits a dataset-style prefix, RL does not directly force that prefix into the
policy. KL penalties, clipping, and reference models can further constrain the
update, but the more basic locality comes from the state source itself.

This helps explain why RL can preserve capabilities even when the reward is
weaker than a full supervised answer. The reward signal may only say whether a
trajectory succeeded, but the trajectory was sampled from the current model. The
update therefore acts on states that are already reachable, making the change a
local policy improvement rather than a global imitation of an external
trajectory distribution.

\subsection{OPD: Teacher-Guided On-Policy Learning}

In OPD, the student samples states and the teacher provides supervision:
\begin{equation}
    \mathcal{L}_{\mathrm{OPD}}(\theta)
    =
    \mathbb{E}_{s \sim d^{\pi_S}}
    \left[
    D(\pi_T(\cdot|s) \,\|\, \pi_S(\cdot|s))
    \right].
\end{equation}
In our strongest OPD variant, the teacher generates short continuations from
student states and the student learns those continuations with cross entropy.
This is analogous to DAgger-style learning: the learner controls the state
distribution, while an expert-like source provides local repair signals
\citep{ross2011reduction}.

OPD is useful because it decouples two properties that are often bundled
together. It keeps the dense supervision of distillation, but moves the state
source from the teacher or dataset to the student. This makes OPD an on-policy
method in the sense that the student decides which prefixes need guidance. The
teacher is not copied as a complete trajectory generator; it is queried as a
local conditional policy on student states.

This distinction explains how a student can surpass a teacher. A teacher's
measured performance depends on both its local conditional distributions and the
states it tends to visit. If the teacher has learned useful local repairs but
also visits poor trajectories, an OPD student can benefit from the repairs
without fully inheriting the teacher's trajectory distribution. The student can
ask, in effect: ``given where I am, what would the teacher do next?'' rather
than ``which states would the teacher have visited instead of me?''

There is also a practical lesson. One-step next-token KL may be too weak for
reasoning tasks because it supplies only a local distribution at each prefix and
does not teach how to complete a trajectory. In our experiments, one-step OPD
collapsed. Continuation-based OPD, where the teacher provides short rollouts
from student states, gives denser trajectory-level supervision while preserving
the on-policy state source.

\section{A Unified Framework: Post-Training as State-Conditioned Supervision}

We view post-training as repeatedly transforming a model's state distribution:
\begin{equation}
    d_{k+1}(s) = \mathcal{T}\big(d_k(s), \mathrm{signal}\big).
\end{equation}
Different algorithms vary both in the source of states and the source of signal.

More explicitly, a post-training step can be written as three operations:
\begin{align}
    s &\sim q_k(s), \\
    z &\sim \mathcal{S}(s), \\
    \theta_{k+1} &=
    \theta_k - \eta \nabla_\theta
    \ell\big(\pi_\theta(\cdot|s), z\big),
\end{align}
where $q_k$ is the training state distribution, $\mathcal{S}$ is the signal
provider, and $z$ is the supervision object: a token, continuation, reward,
preference, or distribution. The resulting policy $\pi_{\theta_{k+1}}$ then
induces a new rollout distribution $d^{\pi_{\theta_{k+1}}}$. The algorithmic
choice of $q_k$ is therefore central. In SFT, $q_k=d_{\mathrm{data}}$ and does
not depend on the current learner. In RL, $q_k=d^{\pi_{\theta_k}}$. In OPD,
$q_k=d^{\pi_{S,k}}$ even though the signal comes from $\pi_T$.

This formulation separates four questions that are often conflated:
\begin{enumerate}[leftmargin=*]
    \item \textbf{Where are updates applied?} This is determined by the state
    source $q_k$.
    \item \textbf{What information is provided?} This is determined by the
    signal source $\mathcal{S}$.
    \item \textbf{How dense is the signal?} Token labels and continuations are
    dense; exact-answer rewards are sparse.
    \item \textbf{How far can the policy move?} This is controlled by learning
    rate, adapter rank, KL penalties, clipping, and optimization details.
\end{enumerate}
Our claim concerns primarily the first question. Objective design matters, but
it is incomplete without specifying the state distribution on which the
objective is evaluated.

\begin{table}[t]
\centering
\caption{State-source view of common post-training methods.}
\label{tab:framework}
\begin{tabular}{lll}
\toprule
Method & Training state source & Supervision signal \\
\midrule
SFT & Dataset trajectories & Gold tokens \\
Offline KD & Teacher trajectories & Teacher logits/tokens \\
OPD & Student trajectories & Teacher logits/continuations \\
RL & Current policy trajectories & Reward \\
DAgger & Learner trajectories & Expert actions \\
\bottomrule
\end{tabular}
\end{table}

This view predicts that methods using learner-induced states can behave
differently from off-policy imitation even when the supervision source is similar
or weaker. In particular, a student can outperform its teacher if the teacher's
errors are coupled to the teacher's own state distribution rather than fully
encoded in its local responses on the student's states.

\subsection{Predictions}

The framework yields several qualitative predictions.

\paragraph{P1: Off-policy pressure can create forgetting under stress.}
When dense supervised updates are applied repeatedly on a narrow external state
distribution, the model may move away from general-purpose behavior. This does
not imply that SFT always forgets; mild SFT can be stable when the dataset
states are compatible with the base policy. The prediction is conditional:
forgetting should appear when off-policy pressure becomes strong enough or
misaligned enough.

\paragraph{P2: On-policy methods should preserve locality.}
RL and OPD should often retain capabilities better than stressed off-policy
training because they apply updates on learner-induced states. This does not
guarantee low scalar drift under every metric, but it predicts that updates are
more likely to be relevant to states the model can actually reach and repair.

\paragraph{P3: Students can surpass teachers.}
If the teacher's failures are partly trajectory-distribution failures, then a
student trained on its own states can outperform the teacher. OPD should be most
effective when the teacher still provides useful local guidance but has degraded
global rollout behavior.

\paragraph{P4: Scalar drift is insufficient.}
Distribution distance between base and post-trained rollouts should be useful,
but it cannot fully characterize training dynamics. Two methods can produce
similar measured drift while differing in which states received supervision and
how locally recoverable those states were. Thus drift should be interpreted
together with state source and signal density.

\section{Experimental Setup}

\paragraph{Model and hardware.}
All experiments use Qwen3-0.6B-Base with LoRA adapters \citep{hu2022lora} on a
single RTX 3090 24GB GPU. We use plain GSM8K-style prompts rather than chat
templates because the model is a base model.

\paragraph{Target and retention tasks.}
The target task is GSM8K \citep{cobbe2021training}. Retention is measured with
TruthfulQA multiple-choice \citep{lin2022truthfulqa} and a selected MMLU subset
\citep{hendrycks2020measuring}. Base model scores are GSM8K 0.448, TruthfulQA 0.300,
and MMLU 0.436.

\paragraph{Methods.}
We evaluate mild SFT, stress SFT, OPD variants, and lightweight on-policy RL. The
mild SFT run uses GSM8K SFT data and produces a non-degraded teacher. The stress
SFT run uses five epochs, learning rate $5\times10^{-4}$, LoRA rank 64, and LoRA
alpha 128, intentionally probing forgetting. OPD samples student states and
trains from teacher continuations. RL uses group-relative exact-answer reward on
GSM8K rollouts.

\paragraph{Metrics.}
For a retention task, forgetting is
\begin{equation}
    F = \mathrm{Score}_{\mathrm{base}} - \mathrm{Score}_{\mathrm{post}}.
\end{equation}
Retention ratio is $\mathrm{Score}_{\mathrm{post}} /
\mathrm{Score}_{\mathrm{base}}$. We report mean forgetting and mean retention
over TruthfulQA and MMLU.

\paragraph{State drift.}
For each trained model, we sample rollouts on a fixed prompt set and collect
prefix states $s_t=(x,y_{<t})$. We embed states with a lightweight lexical
feature representation and report maximum mean discrepancy (MMD) with an RBF
kernel \citep{gretton2012kernel}. We also compute centroid distance, sliced
Wasserstein distance \citep{rabin2011wasserstein}, and lexical Jaccard distance,
but use MMD as the primary scalar drift measure.

\section{Results}

\subsection{SFT Can Be Gentle or Destructive}

\begin{table}[t]
\centering
\caption{Main results. GSM8K is the target task; TruthfulQA and MMLU are
retention tasks. Forgetting and retention are averaged over TruthfulQA and
MMLU.}
\label{tab:main-results}
\begin{tabular}{lrrrrrr}
\toprule
Run & GSM8K & TruthfulQA & MMLU & MMD & Forgetting & Retention \\
\midrule
Base & 0.448 & 0.300 & 0.436 & -- & -- & -- \\
Mild SFT & 0.512 & 0.295 & 0.444 & 0.00956 & -0.0015 & 1.0008 \\
Stress SFT & 0.420 & 0.245 & 0.364 & 0.01093 & 0.0635 & 0.8258 \\
OPD from mild SFT & 0.512 & 0.290 & 0.434 & 0.01470 & 0.0060 & 0.9810 \\
OPD from stress SFT & 0.466 & 0.275 & 0.430 & 0.01092 & 0.0155 & 0.9515 \\
On-policy RL & 0.472 & 0.290 & 0.442 & 0.01098 & 0.0020 & 0.9902 \\
\bottomrule
\end{tabular}
\end{table}

Mild SFT improves GSM8K from 0.448 to 0.512 with essentially no retention loss.
This is an important negative control: SFT does not necessarily forget in our
setup. However, stress SFT produces substantial retention degradation:
TruthfulQA falls from 0.300 to 0.245 and MMLU falls from 0.436 to 0.364. Its
mean retention ratio is 0.8258. Interestingly, stress SFT also reduces GSM8K to
0.420, indicating that overly aggressive off-policy training can degrade both
general and target behavior.

\subsection{OPD Can Surpass a Degraded Teacher}

The clearest OPD result uses the stress SFT model as the teacher. The teacher is
degraded: GSM8K 0.420, TruthfulQA 0.245, MMLU 0.364. OPD from this teacher
achieves GSM8K 0.466, TruthfulQA 0.275, and MMLU 0.430. Thus the student
surpasses its teacher on all measured tasks despite using that teacher as its
supervision source.

This supports the claim that teacher behavior is not transferred as a single
global object. The student receives local guidance on states sampled from the
student's own policy. When those states differ from the teacher's problematic
trajectory distribution, the student can avoid inheriting some teacher failures.

\subsection{RL Provides the On-Policy Reward Point}

The lightweight RL run improves GSM8K from 0.448 to 0.472 while retaining
TruthfulQA 0.290 and MMLU 0.442. Its mean forgetting is only 0.0020. This is
consistent with the view that RL behaves as an on-policy local improvement
method: it changes behavior where the policy samples states and does not require
large off-policy movement.

\subsection{Drift Magnitude Is Not the Whole Story}

Our initial hypothesis was that scalar state drift would strongly explain
forgetting. The data suggest a more nuanced conclusion. Stress SFT and OPD from
the stress teacher have nearly identical MMD drift, 0.01093 and 0.01092, but very
different retention ratios, 0.8258 and 0.9515. Similarly, RL has comparable MMD
drift, 0.01098, with much smaller forgetting.

Therefore, the evidence does not support a simple scalar law of the form
``larger MMD implies more forgetting'' in this small setup. Instead, it supports
a state-source claim: the \emph{quality, locality, and learner-dependence} of
training states matter. Measuring only the distance between rollout
distributions can miss whether updates were applied on states that are locally
recoverable for the learner.

\section{Discussion}

\paragraph{Objective-level analysis is incomplete.}
The contrast between stress SFT and OPD from the stress teacher is difficult to
explain using only the source of supervision. Both are ultimately shaped by the
same degraded teacher/data behavior, but OPD applies supervision on states
sampled from the student. This changes the learning problem.

\paragraph{On-policy dense shaping.}
The successful OPD variant used teacher continuations rather than one-step
logit matching. One-step OPD collapsed badly in our runs, reaching GSM8K 0.040.
Continuation-based OPD recovered target performance. This suggests a practical
recipe: combine on-policy sampling with dense, trajectory-level local
supervision.

\paragraph{Limitations.}
This study is small-scale: one base model, one target dataset, LoRA adapters,
limited retention tasks, and lightweight drift estimators. The RL trainer is a
minimal on-policy GRPO-style implementation rather than a full-scale verl PPO or
GRPO setup. Our drift metric uses lexical features rather than hidden-state or
encoder embeddings. The results should therefore be read as evidence for a
mechanistic hypothesis, not as a benchmark claim.

\section{Related Work}

\paragraph{Post-training objectives.}
Language model post-training is commonly described through the optimization
objective being used. SFT is typically framed as maximum-likelihood learning on
instruction demonstrations. RLHF and related approaches optimize model samples
against learned or verifiable rewards
\citep{christiano2017deep,ziegler2019fine,stiennon2020learning,ouyang2022training}.
Policy-gradient post-training commonly builds on PPO-style conservative policy
optimization \citep{schulman2017proximal}, while recent reasoning systems also
use verifiable rewards and group-relative updates \citep{shao2024deepseekmath}.
Preference-optimization methods such as DPO remove the explicit online RL loop
and express preference learning as a supervised objective
\citep{rafailov2023direct}; related work studies broader families of
preference-optimization losses \citep{azar2024general}. This objective-centric
view has clarified many algorithmic trade-offs, but it can obscure the role of
the state distribution on which the objective is applied. Our work keeps the
objective visible, but treats the training state source as a separate axis.

\paragraph{Exposure bias and imitation learning.}
The distinction between dataset states and learner-induced states has a long
history in imitation learning. Behavioral cloning trains on expert trajectories
and can suffer from compounding errors when the learner visits states absent
from the demonstrations. DAgger addresses this by collecting learner-induced
states and querying an expert on those states \citep{ross2011reduction}. Exposure
bias in sequence prediction captures a related mismatch between training on
gold prefixes and testing on model-generated prefixes
\citep{bengio2015scheduled,ranzato2015sequence}. We adapt this perspective to
LLM post-training: SFT resembles behavioral cloning on fixed trajectories,
whereas RL and OPD apply signals on states induced by the current learner.

\paragraph{Knowledge distillation.}
Knowledge distillation transfers behavior from a teacher to a student through
soft targets, logits, or generated data
\citep{buciluǎ2006model,hinton2015distilling}. Sequence-level distillation and
compressed language-model distillation show that generated teacher outputs can
be effective training data \citep{kim2016sequence,sanh2019distilbert}, and
surveys organize many variants by the knowledge representation and divergence
being optimized \citep{gou2021knowledge}. Our OPD experiments isolate a
different factor: the teacher may provide the signal, but the student can
control the state distribution. This explains why a student need not inherit all
failures of a degraded teacher, especially when teacher supervision is queried
on student states rather than copied from teacher trajectories.

\paragraph{Catastrophic forgetting and retention.}
Catastrophic forgetting has been studied in continual learning as the loss of
previous capabilities when training on new tasks
\citep{mccloskey1989catastrophic,french1999catastrophic,kirkpatrick2017overcoming,lopez2017gradient}.
In LLM post-training, forgetting is often measured as degradation on broad
retention tasks after specialization. Our experiments follow this empirical
tradition by measuring TruthfulQA and MMLU after GSM8K post-training. The
results suggest that forgetting is not only a matter of update magnitude or
dataset size: aggressive off-policy SFT can damage retention, while on-policy RL
and OPD can preserve more capability under comparable target-task pressure.

\paragraph{State drift measurement.}
Distribution shift is often quantified using embedding distances, classifier
two-sample tests, MMD \citep{gretton2012kernel}, or Wasserstein-style metrics
\citep{rabin2011wasserstein,arjovsky2017wasserstein}. We use rollout-state MMD
as a compact proxy for state drift, while also tracking other lexical
distribution statistics. Our results show both the value and limits of such
scalar measures: stress SFT and OPD from the stress teacher have nearly
identical MMD drift, but very different retention. This motivates richer
state-distribution analyses that consider not only how far model rollouts move,
but which training states receive supervision and whether they are locally
reachable by the learner.

\section{Conclusion}

We proposed a state-distribution view of post-training. In this view, SFT is
off-policy state fitting, RL is on-policy reward-guided improvement, and OPD is
teacher-guided on-policy learning. Our experiments show that OPD can surpass a
degraded SFT teacher and that on-policy RL improves a target task with little
retention loss. The evidence refines the original thesis: post-training is not
only about token objectives or scalar distribution drift; it is about where
supervision is applied in the model's state space.

\bibliographystyle{plainnat}
\bibliography{references}

\end{document}